# Performance Analysis and Comparison of Machine and Deep Learning Algorithms for IoT Data Classification

Meysam Vakili[1], Mohammad Ghamsari[2] and Masoumeh Rezaei[3]

1 Department of Computer Engineering, University of Science and Culture, Tehran, Iran
2 Department of Electrical Engineering, Sharif University of Technology, Tehran, Iran
3 Department of Computer Engineering, Khorasan Institute of Higher Education, Khorasan, Iran

**Abstract**

**In recent years, the growth of Internet of Things (IoT) as an emerging technology has been unbelievable. The number of network-enabled devices in IoT domains is increasing dramatically, leading to the massive production of electronic data. These data contain valuable information which can be used in various areas, such as science, industry, business and even social life. To extract and analyze this information and make IoT systems smart, the only choice is entering artificial intelligence (AI) world and leveraging the power of machine learning and deep learning techniques. This paper evaluates the performance of 11 popular machine and deep learning algorithms for classification task using six IoT-related datasets. These algorithms are compared according to several performance evaluation metrics including precision, recall, f1-score, accuracy, execution time, ROC-AUC score and confusion matrix. A specific experiment is also conducted to assess the convergence speed of developed models. The comprehensive experiments indicated that, considering all performance metrics, Random Forests performed better than other machine learning models, while among deep learning models, ANN and CNN achieved more interesting results.**

## 1. Introduction

Internet of Things (IoT) [1] has captured significant attention and investment from both academia and industry in the past few years. Owing to the extremely rapid growth of IoT networks and connected devices to the Internet, this concept has obtained immense popularity in various domains, such as smart houses, smart healthcare, smart transport, smart farming, etc. Different definitions have been proposed for IoT so far, but according to European Research Cluster on the Internet of Things (IERC) IoT is "*A dynamic global network infrastructure with self-configuring capabilities based on standard and interoperable communication protocols where physical and virtual "things" have identities, physical attributes, and virtual personalities and use intelligent interfaces, and are seamlessly integrated into the information network*" [2].

The ubiquity of IoT systems and IoT devices has led to the production of enormous amount of data with specific features known as "Big Data" [3] today. It is obvious that these data are just raw data which are almost meaningless and need some technical methods to gain information from. Thus, we require data processing and analysis techniques to extract useful and meaningful information from the big data generated by IoT devices. These data can also be used for making predictions to better manage possible situations in the future. Furthermore, one of the most significant issues in the IoT world is the smartness of IoT systems, which means moving from traditional IoT to intelligent IoT. To address all of the above-mentioned challenges, the best option is probably getting help from artificial intelligence (AI), and especially machine learning, as the most noticeable part of AI [4]. With proposing a wide range of powerful algorithms, machine learning is capable of processing and analyzing data and extracting beneficial information which is not easily detectable by humans in most cases. In addition, ability of these algorithms in discovering hidden patterns in data, finding semantic relations among data points, and predicting future data using historical data, can make IoT systems smarter and much more efficient in decision making.

Considering above discussion, the point at issue is whether machine learning algorithms can perform good enough in face with Internet of Things data or not. Many researchers in their works have used machine learning for data analysis and future events prediction. Recently, with the emergence of IoT, the challenge of using machine learning for solving IoT problems has become much more popular with researchers. However, these studies are in their infancy and extensive research is still needed to be done. Most presented papers working on the integration of IoT and machine learning have just focused on performance of machine learning algorithms for a specific type of problems. However, in this paper, our purpose is providing a more comprehensive study in order to analyze and compare the performance of several common machine and deep learning [5] algorithms. To achieve this goal, we will use several IoT datasets that come from various domains and each one has different dimensions with different features. For better analysis of results, different performance evaluation metrics will be employed and comparative tables and diagrams will be presented. We specifically concentrate on classification problem which is of utmost importance in the large number of IoT applications.

The rest of the paper is organized as follows. Section 2 will discuss the related work in the area of using machine learning algorithms in IoT domains. In section 3, the methodology will be proposed and detailed information about used algorithms, datasets and performance evaluation metrics will be provided. Sections 4 will investigate and compare the performance of algorithms and finally, in section 5, concluding points will be presented.

## 2. Related Work

In this section, we briefly review some studies that are most relevant to our work. At the end of this section, table 1 summarizes all reviewed articles in terms of their main topics, used algorithms and evaluation metrics. As mentioned in the previous section, the main goal of the most papers presented here is leveraging machine learning methods to solve a particular problem. According to our exhaustive studies, not only are there very few number of research works concentrating on the performance analysis of machine and deep learning algorithms on different types of IoT problems, but they are not also really deep and thorough.

Paper [6] investigated the performance of five popular supervised machine learning algorithms on five different IoT datasets. Among all of the algorithms, Decision Trees achieved the highest accuracy of 99% for all datasets, whereas Logistic Regression and Naïve Bayesian showed the weakest performances. Authors in [7] exploited four machine learning algorithms with the aim of predicting air-conditioning load in a shopping mall. Moreover, they studied hyperparameters optimization for SVR model. The obtained results showed that Chaos-SVR and WD-SVR which are hybrid models outperform single ones, however the complexity of hybrid models can increase considerably. The main objective of [8] is to compare different classification algorithms for an activity recognition problem. The data was produced by mobile phone's accelerometer sensors and Weka workbench was used for data analysis. Based on the results, IB1 and IBk algorithms from lazy classifier category had the best performance in terms of overall accuracy rate for hand palm's position.

The purpose of Ever et.al in [9] is to determine optimum method for prediction problems using nine datasets and four machine learning algorithms. They concluded that the increase or decrease in the number of dataset samples do not influence the performance of algorithms directly, while the features of dataset do. Furthermore, neural network models were able to produce better and more stable results. [10] leveraged a new stochastic algorithm for time series forecast, called the Conditional Restricted Boltzmann Machine (CRBM), in order to predict the energy usage in an office building. Conducted experiments indicated that CRBMs performed better than Artificial Neural Networks and Hidden Markov Models. Research done in [11] compared the performance of a group of machine learning techniques to predict human actions in a smart home using some specific experiments. According to the results, Support Vector Machine gained best accuracy rates over all combinations of input sensors and when the network was equipped with all provided sensors, the most accurate outcomes were achieved.

Authors in [12] evaluated the performance of a wide range of common machine learning algorithms for anomaly detection problem using smart building datasets. They proposed a recommendation framework that can be useful for choosing the most appropriate learning models according to the collected data. It is noticeable that auROC was used as the evaluation metric in this work. Paper [13] employed Linear Regression, Artificial Neural Networks and Support Vector Machines for vehicle classification. Measurements produced by road side sensors including accelerometer and magnetometer have been used in this study. Final results showed that Logistic Regression had the best performance and reached 93.4% for overall classification rate. Wireless based indoor location classification with the help of machine learning algorithms was studied in [14]. This research classified four indoor rooms according to wireless signal strengths from seven various sources. Standard score and feature scaling are normalization methods which were used in this work. According to the experiments, K-Nearest Neighbors had the best performance with around 98% accuracy rate.

Baldominos et.al in [15] compared several well-known machine and deep learning models for human activity recognition by means of generated data from two smart devices: one worn on the wrist and another one in the pocket. Extremely Randomized Trees and Random Forest algorithms showed the best results, using just the sensors positioned on the wrist. Moreover, deep learning models were not able to produce competitive outcomes in the conducted experiments. The main contribution of [16] is the comparison of common machine learning algorithms using an activity recognition android application. This paper performed its experiments on both offline and real-time situations. It has been shown that ANN achieved 97% of recognition rate which was better than all previous experiments without using parameter search. Results also indicated that using just one sensor can lead to generating the same data for different activities, therefore, a single sensor is not always sufficient for getting high recognition rates. The goal of authors in [17] is the analysis of eight data mining and deep learning algorithms on three Internet of Things datasets. According to the experiments, not only did C4.5 and C5.0 obtain better accuracy, but they also had higher processing speeds and consumed memory more efficiently. In [18] a comparative analysis of three supervised algorithms was proposed. Case-Based Reasoner (CBR) reached an accuracy of 92% which was the highest among others. In addition, with respect to the results, when datasets are small the performance of K-Nearest Neighbors is better, while Naïve Bayesian performance stays almost constant even if the size of a dataset and the number of features increase.

## 3. Research Methodology

In this section we are going to describe our approach, especially our selected algorithms, the datasets and their characteristics, and the evaluation metrics that we will use in section 4 for performance comparison.

### 3.1. Machine and Deep Learning Algorithms

Up until now, many machine learning algorithms have been proposed by researchers. Each of these algorithms possess their own strengths and weaknesses which make them suitable for employing in specific domains. More novel versions of these algorithms which have recently gained tremendous popularity, are known as deep learning algorithms. However, the advent of these methods stems from biological neural networks which their introduction to computer science dates back to many years ago. In our study, not only do we use machine learning algorithms, but some of the most common deep learning techniques are chosen to be evaluated on IoT datasets. The performance analysis and comparison of this noticeable number of algorithms on datasets from various IoT domains and with unique characteristics, and the results which will be produced in this thorough investigation, will definitely help researchers and practitioners of related fields to achieve better knowledge and insight for choosing right algorithms according to their problems. Machine learning algorithms used in this paper are Logistic Regression (LR) [19], K-Nearest Neighbors (KNN) [20], Gaussian Naïve Bayesian (GNB) [21], Decision Trees (DT) [22], Random Forests (RF) [23], Support Vector Machine (SVM) [24], Stochastic Gradient Descent Classifier (SGDC) [25] and Adaboost [26]. Moreover, among deep learning algorithms, we analyze the performance of Artificial Neural Networks (ANN) [27], Convolutional Neural Networks (CNN) [5] and Long Short-Term Memory (LSTM) [28] as the most renowned type of recurrent neural networks model.

Table 1. Summary of reviewed research papers

| | Research Paper | Research Topic | Algorithms | Evaluation Metrics |
|---|---|---|---|---|
| 1 | Khadse et.al [6] | Performance comparison of algorithms using IoT data | KNN, NB, DT, RF, LR | Confusion matrix, Accuracy, Precision, Recall, F1-score, Kappa, Execution Time |
| 2 | Xuan et.al [7] | Cooling load forecasting of shopping malls | Chaos-SVR, WD-SVR, SVR, BP | EEP, MBE, $R^2$, Execution time |
| 3 | Ayu et.al [8] | Human activity recognition | NB, SMO, RF, Multilayer Perceptron (MLP), DT, RBFNetwork, IB1, IBK, KStar, Bagging, LogitBoost, etc. | Confusion matrix, Accuracy |
| 4 | Ever et.al [9] | Prediction for different types of problems | BP, RBFNN, SVR, DTR | MSE, Explained Variance (EV), $R^2$ |
| 5 | Mocanu et.al [10] | Energy consumption in buildings | ANN, HMM, CRBM | Correlation Coefficient, RMSE |
| 6 | Alhafidh et.al [11] | Human activity recognition | NB, SVM, RF | Accuracy, Execution Time |
| 7 | Almaguer-Angeles et.al [12] | Anomaly detection using smart buildings data | Bagging, Extra Trees (ET), Gradient boosting, RF, DT | Area under the Receiver Operating Characteristic curve (auROC curve) |
| 8 | Kleyko et.al [13] | Vehicle classification using road side sensors | LR, NNs, SVM | Confusion matrix, Accuracy |
| 9 | Sabanci et.al [14] | WiFi based indoor localization | ANN, KNN, DT, NB, ELM, SVM | Confusion matrix, Accuracy |
| 10 | Baldominos et.al [15] | Human activity recognition | ET, NB, LR, KNN, RF, MLP | Accuracy, F1-score |
| 11 | Suto et.al [16] | Human activity recognition | ANN, CNN | Accuracy |
| 12 | Alam et.al [17] | Performance comparison of algorithms using IoT data | SVM, KNN, LDA, NB, C4.5, C5.0, ANN, DLANN | Confusion matrix, Accuracy, Execution Time |
| 13 | Chettri et.al [18] | Performance comparison of algorithms using IoT data | KNN, NB, CBR | Accuracy, Execution Time |

## 3.2. Datasets

The type of data which we work with, is an influential and determinative issue in selecting algorithms that are best fit for our problems. As it is mentioned earlier, most of the studies conducted on performance evaluation of machine and deep learning methods so far, have focused on a certain type of problem. In addition, just a few number of them have been worked on IoT-related datasets. A valuable contribution of this paper is using different IoT datasets coming from various scopes with distinguished features. All of these datasets belong to smart environments and have been generated by means of IoT-compatible devices and equipment. Six datasets are used in this work which are named DS1 to DS6 for simplicity. In the following, a brief description of all datasets is proposed and a summary of their features is provided in Table 2.

- **DS1** [20]: It is a transportation mode detection dataset collected using smartphones. Transportation mode detection/recognition can be considered as a Human Activity Recognition (HAR) [30] task, but it aims to identify the type of transportation individuals are using. Three different types of sensors have been used for data collection in this dataset, naming accelerometer, gyroscope and sound. There are five target classes including car, bus, train, walking and still.
- **DS2** [31]: This dataset is related to occupancy detection of an office room with the help of four environmental features, namely light, temperature, humidity and $CO_2$. Real-time occupancy detection and estimation are crucial issues in making buildings and indoor places smarter when it comes to energy efficiency. The dataset only focuses on binary classification, which means there are just two target classes or labels (0 and 1) that identify whether the room is occupied by people or not.
- **DS3** [32]: This is an activity recognition dataset containing data from a wearable accelerometer mounted on the chest. The data were collected from 15 participants performing 7 activities. Each class is made up of one activity or a combination of several activities. The classes are: 1) working at computer, 2) standing up, walking and going up/down stairs, 3) standing, 4) walking, 5) going up/down stairs, 6) walking and talking with someone, and 7) talking while standing.
- **DS4** [33]: Activities of daily living (ADL) information related to 14 participants along with their medical records have been collected in this dataset. It has been designed for a fall detection system and has been gathered by means of wearable motion sensors positioned on participants' bodies. Target classes of the dataset include standing, walking, sitting, falling, cramps and running.
- **DS5** [34]: The fifth dataset contains daily weather observations from a large number of weather stations in Australia. It is a huge dataset with values gathered from different weather sensors and measuring equipment. The collected data are about temperature, rainfall, evaporation, sunshine, direction and speed of wind, humidity, pressure and clouds. This dataset is also for a binary classification problem, in which it can be used to predict if it is rainy tomorrow or not.
- **DS6** [35]: This dataset consists of pump sensors data for predictive maintenance. It includes data from 52 sensors which monitor the status of a water pump in a small area, in order to predict the failure time of the pump. The aim of this dataset is to discover whether the water pump is working fine or not. Therefore, this is also related to a binary classification problem.

Table 2. Datasets and their characteristics

| Datasets | Topic | Number of Samples | Number of Features | Number of Target Classes |
|---|---|---|---|---|
| DS1 | Transportation Mode Detection | 5893 | 14 | 5 |
| DS2 | Occupancy Detection | 20560 | 7 | 2 |
| DS3 | Human Activity Recognition | > 1 million | 4 | 7 |
| DS4 | Fall Detection | 16400 | 7 | 6 |
| DS5 | Rain Prediction | 142000 | 24 | 2 |
| DS6 | Pump Failure | 220000 | 54 | 2 |

### 3.3. Performance Evaluation Metrics

After applying machine learning algorithms, we need some tools to find out how well they performed their jobs. These tools are called performance evaluation metrics. A significant number of metrics have been introduced in studies, where each one considers certain aspects of an algorithm performance. Thus, for each machine learning problem we require an appropriate set of metrics for performance evaluation. In this paper, we use several common metrics for classification problems to obtain valuable information about the performance of algorithms and to run a comparative analysis. These metrics are precision, recall [36], f1-score [37], accuracy, confusion matrix and ROC-AUC score [38, 39].

1) **Precision**: It simply shows "what number of selected data items are relevant". In other words, out of the observations that an algorithm has predicted to be positive, how many of them are actually positive. According to formula (1), the precision equals the number of true positives divided by the sum of true positives and false positives:

$$Precision = \frac{TP}{TP+FP} \quad (1)$$

2) **Recall**: It presents "what number of relevant data items are selected". In fact, out of the observations that are actually positive, how many of them have been predicted by the algorithm. According to formula (2), the recall equals the number of true positives divided by the sum of true positives and false negatives:

$$Recall = \frac{TP}{TP+FN} \quad (2)$$

3) **F1-score**: This metric, which is also known as f-score or f-measure, takes both precision and recall into consideration in order to calculate the performance of an algorithm. Mathematically, it is the harmonic mean of precision and recall formulated as follows:

$$F1 - score = 2 \times \frac{Precision \times Recall}{Precision + Recall} \quad (3)$$

4) **Accuracy**: It is the most used and maybe the first choice for evaluating an algorithm performance in classification problems. It can be defined as the ratio of accurately classified data items to the total number of observations (formula (4)). Despite the widespread usability, accuracy is not the most appropriate performance metric in some situations, especially in the cases where target variable classes in the dataset are unbalanced.

$$Accuracy = \frac{TP + TN}{TP + TN + FP + FN} \quad (4)$$

5) **Confusion Matrix**: This matrix is one of the most intuitive and descriptive metrics used to find the accuracy and correctness of a machine learning algorithm. Its main usage is in classification problems where the output can contain two or more types of classes. For more information see [40].

6) **ROC-AUC score**: This metric is calculated using ROC curve (receiver operating characteristic curve) which represents the relation between true positive rate (aka sensitivity or recall) and false positive rate (1- specificity). Area Under ROC Curve or ROC-AUC is used for binary classification and demonstrates how good a model is in discriminating positive and negative target classes. Especially, if the importance of positive and negative classes are equal for us, ROC-AUC score can be a useful performance metric.

## 4. Experiments and Results

In this section, we are going to evaluate and compare the performance of aforementioned algorithms on 6 datasets. Experiments are divided into two parts. In the first part, the algorithms will be analyzed using evaluation metrics including precision, recall, f1-score, accuracy (for both training and test sets) and execution time. For binary classification problems, we also benefit from ROC-AUC score. In the second part of our experiments, we will conduct a more challenging analysis and will compare the proposed algorithms in terms of convergence speed on each dataset.

For each algorithm, we implemented many models with different configurations in order to obtain the best results for each dataset. Moreover, for calculating the average of test set accuracy, we employed 10-fold cross validation. It is important to note that in preparation of machine and deep learning models, we tried our best to achieve a proper balance between the values of performance metrics, especially accuracy, and models execution times. In fact, for some models, deep models to be exact, we obtained high accuracies at the expense of higher execution time. However, when we work on IoT problems, processing time is of utmost importance in evaluation of an IoT system performance. As a consequence, for IoT systems and devices, we are not allowed to develop models which are extremely complex and need a lot of time-consuming computations. On the other hand, we should not forget that most IoT devices are resource and energy-constrained. Thus, the models that have been developed in this study do not have high execution time, but at the same time are able to produce good results.

Although implemented machine and deep learning algorithms for each dataset are compered in one table, we will discuss some interesting points about deep models in particular. We should note that for datasets number 3 and 5, we omitted SVM from our tables, in that its execution time exceeded our threshold immensely. In the tables, the highest figures for each evaluation metric appear in boldface.

All the experiments were conducted on a system with an Intel Corei7 CPU at 2.4GHz, 8 GB of RAM, 1 TB of secondary storage, running windows 10 and Python 3.7. We used Spyder IDE for model development and leveraged machine and deep learning libraries and packages in scikit-learn, TensorFlow and Keras.

## 4.1. Experiment 1: Performance Comparison using Evaluation Metrics

**1) DS1**: This dataset is a completely balanced dataset, meaning for each target class there are equal number of labels. Therefore, accuracy rate can be a quite reliable performance metric for evaluating learning models. According to table 3, in general, RF had the best performance and obtained better results for almost all metrics. The average accuracy rate of RF for test set is 85% and its highest accuracy is 87%. However, the comparison of training and test set accuracies indicates that RF and especially DT suffer from overfitting problem. In terms of execution time, GNB was the fastest, followed by KNN. But it is clear that GNB had the worst performance among other algorithms. As it was predictable, deep learning algorithms, naming ANN, CNN and LSTM, considering their structures, had the highest execution time and LSTM was the slowest. One outstanding point is that CNN performed very well and was much faster than ANN and LSTM.

**2) DS2:** This dataset is related to a binary classification problem. We should mention that it has a feature named "light" which can highly bias the results. In other words, by just using this feature we can achieve 99% accuracy. Owing to this, we eliminated this feature from the original dataset so as to make our analysis much more interesting and challenging. As we can see in table 4, there are several algorithms which demonstrated outstanding performance for this dataset. For precision and f1-score, KNN and RF had the highest values with 98%. For recall, SVM together with these two algorithms obtained the best results and peaked at 98%. The highest figures for test set accuracy belong to KNN and RF, whereas DT and RF gained complete accuracy for training set. When it comes to ROC-AUC score, KNN, RF and SVM were leading algorithms with 0.98 out of 1. Again, GNB had the lowest execution time. Among deep learning algorithms, the performance of ANN and CNN where better, but their execution times are not comparable with those of machine learning models.

Table 3. Performance comparison of the algorithms on DS1

| Algorithms | Precision (%) | Recall (%) | F1-Score (%) | Training Set Accuracy (%) | Test Set Accuracy (%) (Avg/Highest) | Execution Time (Sec) |
|---|---|---|---|---|---|---|
| LR | 63 | 62 | 62 | 64 | 63/65 | 0.68 |
| GNB | 54 | 51 | 48 | 54 | 53/58 | **0.003** |
| KNN | 83 | 83 | 83 | 87 | 80/82 | 0.005 |
| DT | 80 | 79 | 79 | **100** | 76/79 | 0.067 |
| RF | **87** | **87** | **87** | **100** | **85/87** | 0.71 |
| SVM | 80 | 80 | 80 | 88 | 79/81 | 9.9 |
| SGDC | 58 | 59 | 57 | 58 | 55/60 | 0.11 |
| Adaboost | 73 | 72 | 72 | 70 | 67/73 | 0.94 |
| ANN | 77 | 76 | 77 | 77 | 75/82 | 28.84 |
| CNN | 82 | 80 | 80 | 83 | 82/82 | 16.03 |
| LSTM | 77 | 76 | 76 | 75 | 73/77 | 43.35 |

Table 4. Performance comparison of the algorithms on DS2

| Algorithms | Precision (%) | Recall (%) | F1-Score (%) | Training Set Accuracy (%) | Test Set Accuracy (%) (Avg/Highest) | ROC-AUC Score | Execution Time (Sec) |
|---|---|---|---|---|---|---|---|
| LR | 79 | 85 | 81 | 84 | 84/85 | 0.85 | 0.11 |
| GNB | 78 | 78 | 78 | 83 | 83/84 | 0.78 | **0.005** |
| KNN | **98** | **98** | **98** | 99 | **98/99** | **0.98** | 0.01 |
| DT | 97 | 97 | 97 | **100** | 98/98 | 0.97 | 0.06 |
| RF | **98** | **98** | **98** | **100** | 98/99 | **0.98** | 1.1 |
| SVM | 96 | **98** | 97 | 91 | 97/98 | **0.98** | 2.6 |
| SGDC | 80 | 83 | 81 | 85 | 84/86 | 0.83 | 0.08 |
| Adaboost | 89 | 85 | 87 | 92 | 91/92 | 0.85 | 1.05 |
| ANN | 94 | 95 | 94 | 96 | 96/96 | 0.95 | 24.52 |
| CNN | 94 | 95 | 95 | 96 | 94/96 | 0.95 | 36.06 |
| LSTM | 87 | 85 | 86 | 90 | 89/90 | 0.85 | 52.22 |

**3) DS3:** This dataset possesses the greatest number of samples among other datasets in this study. With respect to table 5, RF and KNN reached 92% accuracy for training set, which was the highest number. The best results for recall and f1-score belong to KNN, while RF overtook KNN a little bit in precision metric. It is noticeable that GNB is the only algorithm which processed all data in less than a second. Another remarkable point is that, for this dataset, LR produced the worst results for almost all metrics and among machine learning models, it was the slowest with 17.11 seconds. Deep models obtained quite similar results and their test set accuracies were around 75%, but ANN execution time was much lower than CNN and LSTM.

**4) DS4:** Looking at table 6, it can be seen that RF was the top-performing algorithm in most cases except in execution time. This algorithm achieved an accuracy of 79% for test set which was more than accuracy rates gained by other algorithms. Moreover, RF processed all data in 1.2 seconds which is an acceptable amount of time. The percentage of training set accuracy for RF and DT was 100, which means these algorithms are overfitted. NB was the most inefficient algorithm on this dataset and its accuracy rate never exceeded 16%. CNN came first among deep learning models, while its execution time with 33.5 seconds was better than ANN and LSTM.

Table 5. Performance comparison of the algorithms on DS3

| Algorithms | Precision (%) | Recall (%) | F1-Score (%) | Training Set Accuracy (%) | Test Set Accuracy (%) (Avg/Highest) | Execution Time (Sec) |
|---|---|---|---|---|---|---|
| LR | 23 | 22 | 17 | 51 | 51/51 | 17.11 |
| GNB | 31 | 41 | 29 | 60 | 60/61 | **0.30** |
| KNN | 84 | **80** | **82** | 93 | **92/92** | 3.24 |
| DT | 83 | 78 | 80 | **95** | 91/91 | 4.8 |
| RF | **85** | 79 | 81 | **95** | **92/92** | 12.1 |
| SGDC | 25 | 24 | 19 | 54 | 48/54 | 6.44 |
| Adaboost | 22 | 24 | 19 | 55 | 55/55 | 13.22 |
| ANN | 59 | 43 | 43 | 76 | 75/76 | 88.7 |
| CNN | 61 | 44 | 45 | 76 | 74/76 | 155.6 |
| LSTM | 45 | 39 | 37 | 77 | 74/77 | 255.44 |

Table 6. Performance comparison of the algorithms on DS4

| Algorithms | Precision (%) | Recall (%) | F1-Score (%) | Training Set Accuracy (%) | Test Set Accuracy (%) (Avg/Highest) | Execution Time (Sec) |
|---|---|---|---|---|---|---|
| LR | 25 | 28 | 26 | 40 | 40/41 | 3.04 |
| GNB | 21 | 23 | 11 | 13 | 14/16 | **0.005** |
| KNN | 67 | 69 | 68 | 76 | 66/68 | 0.009 |
| DT | 70 | 71 | 71 | **100** | 70/73 | 0.07 |
| RF | **77** | **76** | **76** | **100** | **76/79** | 1.2 |
| SVM | 63 | 69 | 65 | 73 | 67/68 | 18.12 |
| SGDC | 26 | 27 | 25 | 32 | 33/39 | 0.22 |
| Adaboost | 47 | 41 | 42 | 45 | 43/47 | 1.2 |
| ANN | 50 | 50 | 48 | 51 | 58/60 | 43.17 |
| CNN | 61 | 60 | 58 | 61 | 60/61 | 33.5 |
| LSTM | 21 | 24 | 22 | 31 | 31/33 | 69.42 |

**5) DS5:** This dataset required more preprocessing tasks, because it has 24 feature, 3 of which are categorical with 16 different values. Therefore, after converting these categorical features to numeric ones using one-hot encoding techniques and getting dummy variables, the number of new dataset features almost tripled and it turned into a huge dataset. In accordance with table 7, ANN was the best algorithm among its competitors and is the only model that reached 86% for test set accuracy. The performance of ANN becomes more valuable when we can see that its ROC-AUC score is 0.74, which was again the highest. For this dataset, the lowest and highest figures for execution time are 0.19 and 188.7 which belong to GNB and LSTM, respectively. It is also noticeable that LR, RF, SGDC, Adaboost and CNN also performed very well and obtained test set accuracies very close to ANN.

**6) DS6:** This is the most voluminous dataset among others with approximately 220,000 samples and more than 50 features. It is extremely unbalanced, therefore, our performance metrics are not able to present meaningful analysis and comparison of machine and deep learning algorithms. Owing to this fact, we benefited from confusion matrices. In table 8, just execution time and confusion matrix of test set for each algorithm are shown. In this problem, the best model is the one that is able to predict the highest number of pump failure; it means, the cell in the confusion matrix, where the number of predicted labels and true labels are both 1. In the confusion matrices of table 8, rows indicate true labels, while columns represent predicted labels. According to this table, 6 algorithms were able to predict all 303 abnormal situations accurately, naming KNN, RF, SVM, Adaboost, ANN and CNN. But among them, RF and Adaboost had the lowest number of wrong predictions. Precisely, there was just one case that these algorithms predicted 1 for target class, whereas the actual value was 0. This competition becomes more interesting when we can observe that RF processed this dataset in only 1.8 seconds, while the figure for Adaboost was 34 seconds. Therefore, the final winner was RF.

Table 7. Performance comparison of the algorithms on DS5

| Algorithms | Precision (%) | Recall (%) | F1-Score (%) | Training Set Accuracy (%) | Test Set Accuracy (%) (Avg/Highest) | ROC-AUC Score | Execution Time (Sec) |
|---|---|---|---|---|---|---|---|
| LR | 72 | 79 | 74 | 85 | 85/85 | 0.72 | 8.2 |
| GNB | 65 | 68 | 65 | 73 | 73/74 | 0.68 | **0.19** |
| KNN | 76 | 72 | 74 | 90 | 81/83 | 0.66 | 10.9 |
| DT | 70 | 71 | 70 | **100** | 79/80 | 0.70 | 2.22 |
| RF | 80 | 71 | 74 | 99 | 85/85 | 0.71 | 1.42 |
| SGDC | 76 | 71 | 73 | 83 | 84/84 | 0.71 | 2.64 |
| Adaboost | 80 | 69 | 72 | 84 | 84/84 | 0.70 | 2.41 |
| ANN | **82** | **74** | **77** | 86 | **85/86** | **0.74** | 35.00 |
| CNN | 78 | 73 | 75 | 85 | 84/85 | 0.73 | 24.8 |
| LSTM | 71 | 59 | 60 | 80 | 78/80 | 0.59 | 188.7 |

Table 8. Performance comparison of the algorithms on DS6

| Algorithms | Execution Time (Sec) | Test Set Confusion Matrix | | | Algorithms | Execution Time (Sec) | Test Set Confusion Matrix | | |
|---|---|---|---|---|---|---|---|---|---|
| | | | 0 | 1 | | | | 0 | 1 |
| LR | 5.74 | 0 | 11598 | 10 | SGDC | **0.71** | 0 | 11589 | 19 |
| | | 1 | 12 | 291 | | | 1 | 10 | 293 |
| GNB | 0.18 | 0 | 11429 | 179 | Adaboost | 34.01 | 0 | **11607** | **1** |
| | | 1 | 2 | 301 | | | 1 | **0** | **303** |
| KNN | 0.73 | 0 | **11606** | **2** | ANN | 20.3 | 0 | **11602** | **6** |
| | | 1 | **0** | **303** | | | 1 | **0** | **303** |
| DT | 3.38 | 0 | 11607 | 1 | CNN | 43.91 | 0 | **11603** | **5** |
| | | 1 | 3 | 300 | | | 1 | **0** | **303** |
| RF | 1.8 | 0 | **11607** | **1** | LSTM | 141.9 | 0 | 11603 | 5 |
| | | 1 | **0** | **303** | | | 1 | 44 | 259 |
| SVM | 11.65 | 0 | **11605** | **3** | | | | | |
| | | 1 | **0** | **303** | | | | | |

## 4.2. Experiment 2: Convergence Speed

In this part, we conduct a quite different experiment. According to our comprehensive studies, no research paper has carried out similar experiments to ours for performance evaluation of machine and deep learning algorithms in IoT domains. We are going to assess proposed algorithms in terms of convergence speed on various datasets. By convergence speed, we mean how fast an algorithm can learn and as the volume of input training data increase gradually, how accuracy rate of the algorithm will change. Thus, the convergence speed here is completely different from execution time. In other words, the aim of convergence speed experiment is to investigate the relation between the volume of training data given to an algorithm as input, and the test set accuracy rate which the algorithm can gain for that amount of data. For different datasets we consider different volumes of input training data. For datasets number 1, 2 and 4 we use four proportions as input data, which means 5, 10, 30 and 50 percent of total data in the dataset. In other words, we first give algorithms 5% of all data in a dataset as training data and calculate the accuracy of all algorithms on a certain test set. We then increase the volume of input data to 10, 30 and 50 percent and repeat the experiment. For DS5 these proportions are 1, 5, 10 and 30 percent and for DS3 these figures are 0.1, 0.5, 1 and 5 percent. This experiment is only carried out on DS1 to DS5.

The major reason for doing this experiment is the nature of IoT data. We know that data in IoT systems are input and collected gradually. As a result, machine and deep learning algorithms that are supposed to be used in IoT devices should have the capability to learn and predict even with small amount of training data. Table 9 indicates complete information related to this experiment in terms of datasets, volume of input training data, algorithms and the percentage of test set accuracy they achieved. However, to gain better and quicker insight, we will illustrate results using line graphs which show top-performing models.

According to Fig. 1, RF had the best performance in terms of convergence speed. After importing just 5% of total data, it gained 77% accuracy and when the input data was increased to 50%, RF reached the highest level, at 88%. SVM, KNN, DT and CNN also showed good performance on this dataset.

For DS2 several algorithms did their jobs very well and obtained high accuracy quickly. As we can see in Fig. 2, only 5% of total data was enough for KNN, DT, RF and SVM to hit 93% and more accuracy rate. They continued their increasing trend of learning and reached 97-99% at the end. Adaboost, ANN and CNN are other successful algorithms for this dataset.

Table 9. Convergence speed comparison of the algorithms on different Datasets

| DS | Training Data Volume | LR | GNB | KNN | DT | RF | SVM | SGDC | Adaboost | ANN | CNN | LSTM |
|---|---|---|---|---|---|---|---|---|---|---|---|---|
| **DS1** | 5% | 53 | 47 | 57 | 57 | 77 | 70 | 37 | 43 | 50 | 57 | 50 |
|  | 10% | 61 | 56 | 73 | 73 | 75 | 63 | 47 | 53 | 64 | 66 | 59 |
|  | 30% | 66 | 62 | 76 | 73 | 82 | 75 | 59 | 64 | 73 | 63 | 62 |
|  | 50% | 68 | 56 | 81 | 76 | 88 | 80 | 61 | 71 | 79 | 79 | 71 |
| **DS2** | 5% | 81 | 84 | 94 | 94 | 94 | 93 | 85 | 91 | 91 | 92 | 85 |
|  | 10% | 84 | 85 | 90 | 92 | 93 | 93 | 83 | 87 | 89 | 88 | 84 |
|  | 30% | 86 | 83 | 99 | 97 | 98 | 97 | 87 | 92 | 93 | 95 | 85 |
|  | 50% | 85 | 83 | 98 | 98 | 99 | 97 | 85 | 92 | 93 | 91 | 88 |
| **DS3** | 0.1% | 53 | 59 | 77 | 75 | 73 | 73 | 62 | 57 | 51 | 64 | 23 |
|  | 0.5% | 51 | 62 | 76 | 71 | 74 | 69 | 49 | 60 | 64 | 68 | 42 |
|  | 1% | 52 | 61 | 74 | 72 | 75 | 71 | 51 | 52 | 65 | 70 | 56 |
|  | 5% | 51 | 61 | 76 | 75 | 79 | 70 | 50 | 55 | 69 | 73 | 59 |
| **DS4** | 5% | 43 | 18 | 52 | 55 | 61 | 45 | 29 | 46 | 52 | 46 | 21 |
|  | 10% | 43 | 20 | 63 | 66 | 72 | 66 | 39 | 34 | 43 | 46 | 25 |
|  | 30% | 39 | 17 | 68 | 66 | 74 | 67 | 32 | 37 | 52 | 53 | 28 |
|  | 50% | 32 | 12 | 67 | 67 | 74 | 65 | 31 | 36 | 47 | 53 | 31 |
| **DS5** | 1% | 78 | 71 | 77 | 77 | 79 | 85 | 81 | 83 | 84 | 75 | 79 |
|  | 5% | 85 | 72 | 80 | 78 | 85 | 87 | 81 | 85 | 84 | 81 | 80 |
|  | 10% | 85 | 75 | 78 | 78 | 84 | 85 | 81 | 85 | 84 | 82 | 79 |
|  | 30% | 84 | 72 | 79 | 78 | 84 | 84 | 81 | 84 | 85 | 84 | 78 |

Fig. 3 shows that by entering just 0.1% of total data, KNN became the fastest learner with an accuracy of 77%. Its outstanding performance were followed by DT, RF, SVM and CNN. It is interesting to note that when the volume of input data has reached 5%, it was RF which came first with 79% accuracy.

With respect to Fig. 4, RF overtook other algorithms for any volume of input data. Its accuracy started at 61% for 5% of input data and peaked at 74% when the training data increased to 50% of whole dataset. DT, KNN and SVM also produced good results, but they were not able to gain an accuracy rate higher than 67%. One surprising point is that none of deep learning models could achieve acceptable results in face with this dataset and the best accuracy that they could obtain was only 53%.

Looking at table 9, it can be seen that most of the algorithms performed quite well on DS5. Maybe the main reason is that this dataset like DS2 is related to a binary classification problem. According to Fig. 5, when our algorithms trained with 1% of input data, SVM, Adaboost and ANN with 85, 84, and 83, had the highest test set accuracy rates, respectively. However, once the volume of training data increased to 30% of total data, LR, RF and CNN progressed very well and reached an accuracy rate between 84-85%. Overall, our models showed small fluctuations in accuracy rates for this dataset and GNB was the weakest algorithm in comparison with others.

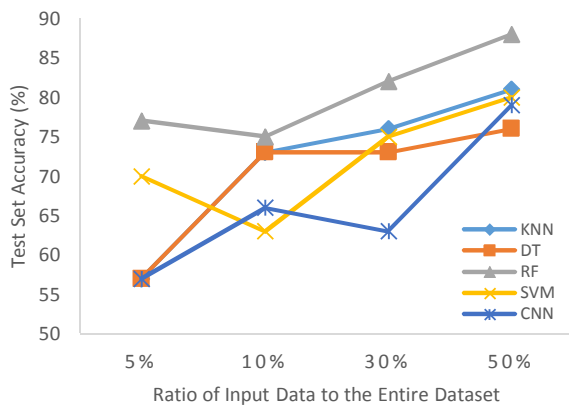

Fig. 1: Convergence speed comparison of top algorithms on DS1

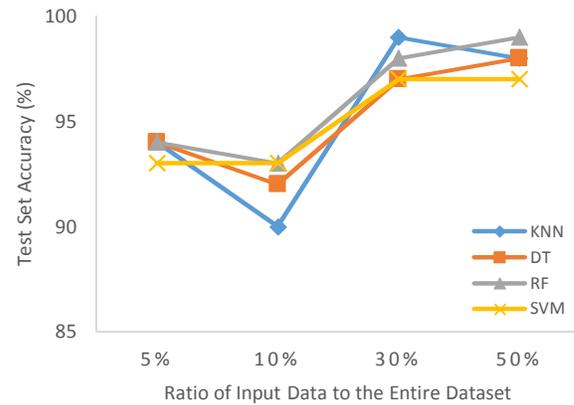

Fig. 2: Convergence speed comparison of top algorithms on DS2

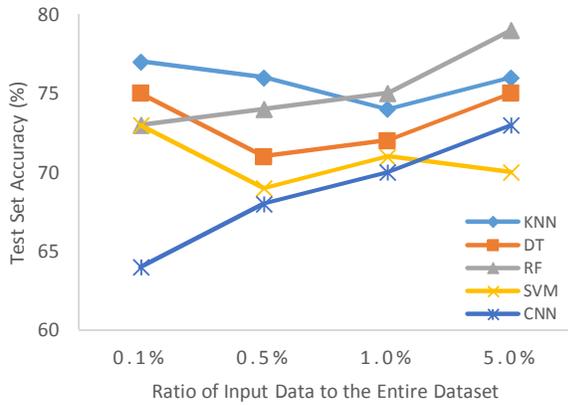

Fig. 3: Convergence speed comparison of top algorithms on DS3

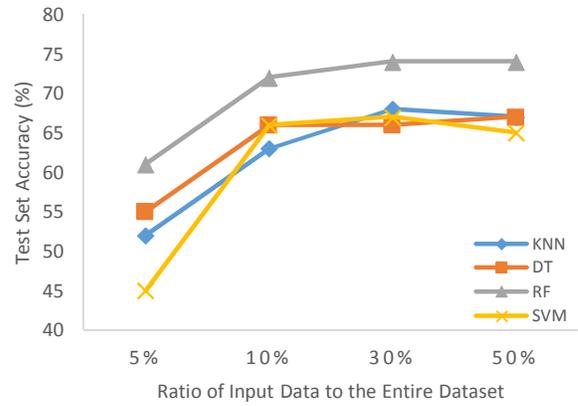

Fig. 4: Convergence speed comparison of top algorithms on DS4

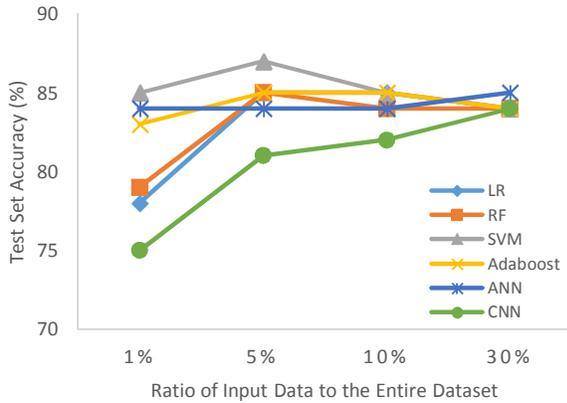

Fig. 5: Convergence speed comparison of top algorithms on DS5

### 4.3. Discussion

Finally, we are going to terminate this section by giving some concluding points on achieved results.

1) LR performed much better for binary classification compared with multi-class classification problems.
2) GNB and SGDC showed weakest performance in most cases.
3) In almost all problems, GNB was the fastest algorithm in terms of execution time.
4) Because of recurrent structure of LTSM, it took the highest execution time for the processing of most datasets. However, for datasets DS3 and DS5, SVM was disappointingly slow and as a consequence, we eliminated its results from our tables.
5) RF and DT were two algorithms which showed higher probability of overfitting.
6) Among deep learning models, CNN produced surprising results. Many research studies use this algorithm mostly for image processing and computer vision problems. But in this study, by conducting different experiments and analyses, we saw that CNN is able to perform very well for typical classification problems. In fact, not only did CNN achieve interesting values for our evaluation metrics, but it also appeared faster in many cases in comparison with ANN and LSTM.
7) In relation to convergence speed experiment, KNN, DT, RF and SVM was the best among machine learning models, while for deep models, ANN and CNN performed better than LSTM.
8) According to our comprehensive studies and experiments, deep models are able to produce extremely good outcomes, if they benefit from a well-designed architecture and near optimal hyperparameter tuning. But the key point is that with a complex architecture, execution time will grow dramatically. Due to this, it may seem that using deep learning algorithms for time-critical IoT problems is not a good idea. On the other hand, we cannot easily ignore the power of these algorithms for processing sophisticated problems. Thus, it is suggested that by leveraging state-of-the-art techniques including distributed and federated learning [41, 42] which spread workloads among processing nodes, we would be able to implement deep learning models on resource and energy-constrained IoT devices.

# 5. Conclusion

In this paper, we carried out comprehensive experiments to evaluate the performance of several machine and deep learning algorithms, naming LR, GNB, KNN, DT, RF, SVM, SGD Classifier, Adaboost, ANN, CNN and LSTM. The distinction between this study and many of other researches is that it targets Internet of Things environments and concentrates on IoT-related datasets for classification problems. We conducted two separated experiments to assess our models. One experiment for investigating the performance of models in terms of some evaluation metrics, and another one for measuring how fast these models can learn. According to the results, RF overtook other machine learning algorithms for most metrics, while the execution time of GNB was lower than others. With respect to deep learning algorithms, they were ANN and CNN which obtained the best results. When it comes to convergence speed, it was again RF which could learn faster when the volume of input training data was remarkably small.